# A Proposed S.C.O.R.E. Evaluation Framework for Large Language Models – Safety, Consensus, Objectivity, Reproducibility and Explainability

Ting Fang Tan[1]

Kabilan Elangovan[1,2]

Jasmine Ong[3,4,5]

Nigam Shah[6]

Joseph Sung[7]

Tien Yin Wong[1,8]

Lan Xue[8]

Nan Liu[9]

Haibo Wang[10]

Chang Fu Kuo[11]

Simon Chesterman[12]

Zee Kin Yeong[13]

Daniel SW Ting[1,2,9]

1. Singapore National Eye Centre, Singapore Eye Research Institute, Singapore, Singapore
2. Singapore Health Services, Artificial Intelligence Office, Singapore
3. Division of Pharmacy, Singapore General Hospital, Singapore
4. Department of Pharmacy, University of California San Francisco, California, USA
5. Duke-NUS Medical School, Singapore, Singapore
6. School of Medicine, Stanford University, Singapore
7. Lee Kong Chian School of Medicine, Nanyang Technological University, Singapore
8. School of Medicine, Tsinghua Medicine, Tsinghua University, Beijing, China
9. Duke-NUS Medical School, Singapore
10. The First Affiliated Hospital, Sun Yat Sen University, China

11. Chang Gung Memorial Hospital, Taiwan
12. Faculty of Law, National University of Singapore, Singapore
13. Singapore Academy of Law, Singapore

**Corresponding author:**
A/Prof Daniel Ting MD (1$^{st}$ Hons) PhD
Associate Professor, Duke-NUS Medical School
Director, AI Office, Singapore Health Service
Head, AI and Digital Health, Singapore Eye Research Institute
Address: The Academia, 20 College Road, Level 6 Discovery Tower, Singapore, 169856

**Abstract**

A comprehensive qualitative evaluation framework for large language models (LLM) in healthcare that expands beyond traditional accuracy and quantitative metrics needed. We propose 5 key aspects for evaluation of LLMs: Safety, Consensus, Objectivity, Reproducibility and Explainability (S.C.O.R.E.). We suggest that S.C.O.R.E. may form the basis for an evaluation framework for future LLM-based models that are safe, reliable, trustworthy, and ethical for healthcare and clinical applications.

**Main**

Since the debut of ChatGPT (generative pre-trained transformer) in 2022, there has been an exponential surge in interest on large language models (LLMs). LLMs utilize advanced deep learning techniques, particularly transformer architectures, to learn complex associations from vast amounts of unstructured text. Unlike traditional neural networks, transformers use attention mechanisms to capture patterns in sequential data, enabling more sophisticated understanding and generation of human language (Figure 1). Generative artificial intelligence (AI) applications built on backend LLMs enable realistic user interactions through text-based dialogue. Studies have shown the feasibility of generative AI in healthcare, demonstrating capabilities such as passing medical board examinations, answering clinical questions, providing medical advice, and interpreting clinical scenarios and investigations. This integrated understanding of AI concepts and their interrelationships underscores the potential of generative AI in advancing healthcare applications[1,2].

While these initial observations suggest the potential of generative AI and specifically LLMs in revolutionizing healthcare delivery, evaluation of these new AI models and their applications have been variable with no consistent metrics employed. Furthermore, observations of 'hallucinations' in generated responses from LLMs, where the models produce content that is entirely fabricated or nonsensical, and 'falsehood mimicry,' where incorrect information is presented in a seemingly accurate and confident manner, are among the cautions raised against LLMs. These issues highlight the importance of critical evaluation and validation when using LLMs in sensitive applications like healthcare. These issues are particularly challenging in clinical medicine and healthcare, where misinformation can result in significant harm to patient safety, such as incorrect diagnoses, prognoses, and clinical management recommendations. Sometimes, the generated information is not only incorrect but also biased, especially regarding controversial topics. Whether it is beneficial to present 'balanced' information is also debatable, as it might confuse patients further[3]. Therefore, there is a compelling need for more detailed and domain-specific evaluation of LLM-driven algorithms in healthcare use cases.
.

To address some of these issues, publicly available benchmark datasets (Eg. PubMedQA[4], MedMCQA[5], MultiMedQA[6], Measuring Massive Multitask Language Understanding (MMLU) clinical topics[7]) have been used to quantitatively compare performance between various LLMs and clinicians of varying expertise. For example, MultiMedQA is a curated dataset of 6 medical question-answering datasets covering medicine, research and consumer queries, and HealthSearchQA is a dataset of online searches of medical questions. These datasets are largely in multiple-choice format, facilitating summation of scores for quantitative, objective and standardized comparison of performance. While standardized test sets based on medical board examinations may not be representative of clinical competency in real-word clinical practice,[8] these datasets nevertheless provide a benchmark and insights from LLM-generated explanations that may serve as an educational tool for residents-in-training in dissecting clinical concepts and supplementing additional resources.

The traditional quantitative metrics employed for LLM evaluation focus on text similarity, where generated responses are compared against a reference text as the ground truth[9,10]. Examples are listed in Table 1[11-14]. These metrics enable quantifiable comparison amongst models that can be efficiently automated. However, while these metrics are useful for traditional NLP tasks such as text summarization or machine translation, they may be less applicable to the healthcare domain. First, they require a reference text as the gold-standard for comparison to generate a score, which may not be as relevant in the clinical setting where there is often no 'model answer' to a clinical question. Next, their focus on exact word or sequence matching may fail to capture the nuances and contextual understanding that are essential in the clinical setting.

Further insight and evaluation into usefulness and application of LLM evaluation may lie in more subjective and qualitative assessment of LLM-generated information. Fundamentally, such types of evaluation are centered on human alignment and specifically the clinician expert as gold-standard. While evaluation based on linguistic features such as fluency and grammar are intrinsic to an LLM, qualitative evaluation to uncover deeper insights beyond answer accuracy and specifically tailored to the healthcare domain needs to be established[17]. Efforts have been made to outline these domain-specific components of evaluation. For example, Articulate Medical Intelligence Explorer (AMIE) is an LLM-based AI system developed by Google for diagnostic medical reasoning via text-based consultations, was evaluated on clinical scenarios with simulated patients in an OSCE format. Evaluation metrics were defined as the accuracy of the top 3 diagnoses, appropriateness of diagnosis and management (10 components), and an emphasis on displaying empathy and addressing concerns via the Practical Assessment of Clinical Examination Skills (PACES) (16 components), and relationship fostering via the Patient-centered Communication Best Practice (PCCBP) (6 components)[18]. MedPALM generated responses were evaluated on a multi-axis framework for human evaluation including alignment with scientific and clinical consensus, likelihood of harm and bias, reading comprehension, recall of relevant clinical knowledge, manipulation of knowledge via valid reasoning, completeness of responses, relevance and helpfulness[6]. The increasing emphasis on incorporating evaluation on LLM trust and safety was similarly echoed in another study that highlighted 7 key categories— including reliability, safety, fairness, resistance to misuse, explainability and reasoning, adherence to social norms, and robustness. Each category is further divided into several sub-categories, with a total of 29 sub-categories[19]. Human grading is time-consuming and labor-intensive[20]. Furthermore, narrative analysis of LLM-generated responses as well as the use of variable evaluation rubrics across studies limit head-to-head comparison amongst applications. These may further impede the development pipeline of newer LLM applications. One example, in line with these objectives, was HELM Instruct proposed by the Stanford University for open-ended, multidimensional, and absolute assessment, encompassing 5 criteria Helpfulness, Understandability, Completeness, Conciseness, and Harmlessness (on a scale of 1-5), targeted at general non-medical open-ended questions. This demonstrated the feasibility and highlighted the importance for multi-dimensional evaluation centered on alignment with domain-experts.

**Methods**

**Proposed S.C.O.R.E. Evaluation Framework**

To allow a more subjective and qualitative assessment of LLMs in healthcare beyond traditional quantitative measures and centered on domain-expert alignment, we proposed the S.C.O.R.E. Evaluation Framework that outlines 5 key aspects of evaluation (Table 2): First, *Safety* is defined as an LLM-generated response not containing hallucinated or misleading content that may lead to physical and/or psychological adversity to the users. Safety includes both accuracy of the LLM tool in offering a diagnosis and recommending intervention that may incur injury to the subject. Ensuring safety involves rigorous testing and validation to prevent the dissemination of false or harmful information, which is crucial in maintaining the integrity and trustworthiness of LLM-based systems in clinical settings. Second, *Consensus* is defined as a response that is accurate and aligned with the clinical evidence and professional consensus according to national and international professional bodies. This alignment with established evidence, medical guidelines and professional expert opinion is essential to ensure that LLM-generated recommendations are credible and reliable. Third, *Objectivity* is defined as a response that is objective and unbiased against any condition, gender, ethnicity, socioeconomic classes and culture. This will help in assessing the responses ethically and ensuring that the LLM provides fair and equitable responses promoting inclusivity and preventing discrimination. Next, *Reproducibility* is defined as a consistent response after repeated response generation to the same question. The focus is not on the word-for-word repeatability of the responses but rather assessing reproducibility in terms of the contextual consistencies between the responses generated. Finally, *Explainability* is defined as justification of the LLM-generated response including the reasoning process and additional supplemental information relevant to the context including reference citations or website links. All responses are graded on Likert Scale from 1 (Strongly disagree) to 5 (Strongly agree). Grading should be conducted by clinical domain experts who have the necessary knowledge and experience to assess the content's relevance and adherence to professional standards. The S.C.O.R.E. Evaluation Framework serves as a broad framework that can be adapted to various disciplines. These components are universally relevant principles that enhance the quality and reliability of outputs from LLM applications across different fields. While the S.C.O.R.E. framework provides a solid foundation, it could be further refined to address the unique challenges and standards of each specialty to maximize its applicability and impact.

Quantitative Metrics against Qualitative S.C.O.R.E Framework

To evaluate the effectiveness of the proposed S.C.O.R.E framework, we conducted head-to-head comparisons using conventional quantitative evaluation metrics, including BLEU, ROUGE-1, ROUGE-L, and BERT-SCORE, to assess LLM-generated open-ended responses to healthcare-related questions. We selected commonly asked patient queries with paired answers related to general ophthalmology and medications. These question-answer (Q&A) pairs were manually crafted by domain experts, with the ophthalmology-related Q&A pairs extracted from a previous study[21]. We utilized GPT4-omni[22] as the LLM for generation of responses, setting the instructional prompt as follows: "You are a medical chatbot interacting with patients regarding their health inquiries. Please provide concise and clinically accurate responses." The hyperparameters were configured with a temperature of 0.2 and a maximum token limit of 256. The temperature was set to 0.2 to ensure more deterministic and focused responses from the model. A lower temperature reduces the randomness in the model's output, leading to more consistent and reliable answers. This is particularly important in clinical and healthcare applications, where precision and consistency are crucial. The 256 token limit helps to prevent overly verbose responses, ensuring that the output remains concise and to the point. The paired answers served as the clinical ground-truth for the quantitative evaluation. Qualitative assessments based on the S.C.O.R.E framework were performed by a board-certified senior consultant ophthalmologist DT and a principal pharmacist JO.

Based on the quantitative evaluation, GPT4-omni responses were deemed suboptimal for both ophthalmology and medication-related queries, as illustrated in Figure 2. The evaluation metrics for ophthalmology-related queries yielded poor average scores: BLEU 0.0238, ROUGE-1 0.2613, ROUGE-L 0.2351, and BERT-SCORE 0.5925. Similarly, for medication-related queries, the scores were BLEU 0.0152, ROUGE-1 0.2484, ROUGE-L 0.2141, and BERT-SCORE 0.5864.

On the other hand, the qualitative assessment using S.C.O.R.E found that the GPT4-omni responses were clinically accurate, as depicted in Figure 3. For ophthalmology-based questions, the average Likert scores were 5 for *Safety*, 4.2 for *Consensus*, 5 for *Objectivity*, 4.6 for *Reproducibility*, and 5 for *Explainability*. Similarly, for medication-based questions, the average Likert scores were 4.8 for *Safety*, 5 for *Consensus*, 5 for *Objectivity*, 4.4 for *Reproducibility*, and 4.4 for *Explainability*. In one of the medication-related question on genetic influence of azathioprine, GPT4-omni's response was aligned with the ground truth in identifying the intended question (adverse drug reaction related to genetic polymorphism), aligning with evidence-based knowledge (genetic polymorphism increases the risk for severe toxicity, guideline-concordant actions) and reinforcing the need for genetic tests prior to initiation (Figure 4). While the response would have otherwise been misrepresented as inaccurate based on the quantitative metrics, the components of S.C.O.R.E. allowed these clinically relevant aspects to be qualitatively assessed. In one of the ophthalmology-related questions on the symptoms of diabetic retinopathy, the GPT4-omni response was notably accurate in highlighting the importance of regular examinations for early detection as symptoms may not be noticeable in early stages. However, the response listed impaired color vision and visual field defects as common symptoms, while not incorrect, these are not typically observed in diabetic retinopathy. Using the S.C.O.R.E. framework, this response was graded a 3 out of 5 for *Consensus*. Therefore, the qualitative assessment facilitated a more nuanced understanding of the clinical accuracy and relevance of LLM-generated responses.

**Integrating with Existing Efforts in Furthering LLM Evaluation**

The objective of S.C.O.R.E is to outline a multi-dimensional framework to facilitate standardized, qualitative and efficient human evaluation of LLM-generated open-ended answers in domain-specific tasks. S.C.O.R.E can potentially be integrated with existing efforts in deepening LLM evaluation. The *Safety* component in S.C.O.R.E. can potentially be expanded to include resilience against adversarial prompting. This is aligned with previous work demonstrating the "willingness" of GPT-3.5 to comply to a harmful prompt in general and medical domains (Eg. falsifying medical records, violating patient confidentiality, and spreading medical misinformation), which was reduced after model fine-tuning with safety demonstrations[23]. Towards deployment, additional layers of assessment such as translational value and governance (Eg. fairness, transparency, trustworthiness, accountability based on the Governance Model for AI in Healthcare (GMAIH) [24]) have also been emphasized[25]. S.C.O.R.E. can potentially be used as the initial evaluation of LLM applications, before further evaluation on translational value and governance. There has also been recent work that explored leveraging LLM-based evaluation, in striving for automated and reference-free evaluation[26]. GPT-4 based evaluation of LLM-generated responses to general ophthalmology-related patient queries were found to be highly congruent with human clinician rankings[27]. This was similarly demonstrated for evaluating general tasks in G-EVAL using chain-of-thought prompting with GPT-4[28]; and in LLM-EVAL using a single-prompt multi-dimensional automatic evaluation of open-domain LLM conversations[29]. Furthermore, the components of the S.C.O.R.E. framework may potentially be embedded into input prompts, to guide LLM generation of high-quality responses that are attuned to critical benchmarks. Nevertheless, domain-expert human evaluation cannot be replaced by LLM-based evaluation, without further work to validate these preliminary observations. Beyond evaluation, future work could also explore LLM enhancement to modify responses based on feedback from LLM evaluation[30].

**Conclusions**

As the capabilities of LLMs continue to expand, effective evaluation beyond traditional quantitative metrics such as accuracy is essential to comprehensively critique these generative AI models and validate them specific to their domain (in this case, healthcare and clinical use). By incorporating factors such as safety, consensus, objectivity, reproducibility and explainability, S.C.O.R.E can ensure that LLM-based models and systems are not only robust and accurate, but are also safe, reliable, ethical, and trustworthy in their application, particularly for clinical medicine and healthcare.

# Tables

## Table 1. Summary of quantitative metrics used for text summarization and machine translation tasks

| | | |
|---|---|---|
| **BLEU** | Bilingual Evaluation Understudy | • Evaluates the quality of LLM-generated translated text by comparing it with a translated reference standard<br>• May be result in brevity penalty<br>• Ranges from 0 to 1, with 1 indicating perfect translation |
| **ROUGE** | Recall-Oriented Understudy for Gisting Evaluation | • Measures the number of n-gram overlap, ie. Consecutive sequences of 'n' number of words, between the LLM generated response and human-generated reference standard<br>• E.g. ROUGE-1 refers to a unigram, ROUGE-2 refers to a bigram, ROUGE-3 refers to a trigram, ROUGE-L (RL) refers to the longest common subsequence |
| **BERTScore** | Bidirectional Encoder Representations from Transformers | • Creates contextual embeddings or numerical representations of words, and measures the distance between associations of words between<br>• More representative comparison of semantic meaning of LLM-generated and reference text |
| **Perplexity** | - | • Measures how well the LLM predicts the a text sample<br>• Lower value indicates better performance. |
| **F1 Score** | 2 x (Precision x Recall) / (Precision + Recall) | • Measures the harmonic mean or the balance between precision (true positives out of all true and false positives) and recall (true positives out of all true positives and false negatives) |
| **METEOR** | Metric for Evaluation of Translation with Explicit ORdering | • Evaluates the quality of LLM-generated translated text by comparing it with a translated reference standard, via unigram matching |

## Table 2. Proposed Evaluation Framework for Domain-specific LLM applications: S.C.O.R.E Evaluation Framework

| S.C.O.R.E. Evaluation Framework | | |
|---|---|---|
| **S**afety | Non-hallucinated responses with no misleading information | **Likert scale 1 to 5**<br>1: Strongly Disagree<br>2: Disagree<br>3: Neutral<br>4: Agree<br>5: Strongly Agree |
| **C**onsensus | Response is accurate and aligned with clinical consensus | |
| **O**bjectivity | Response is objective and unbiased against any condition, device, or demographic | |
| **R**eproducibility | Consistency of responses after repeated generation to the same question | |
| **E**xplainability | Justification of response including reasoning process and additional supplemental information | |

**Figure 1: Evolution of Artificial Intelligence**

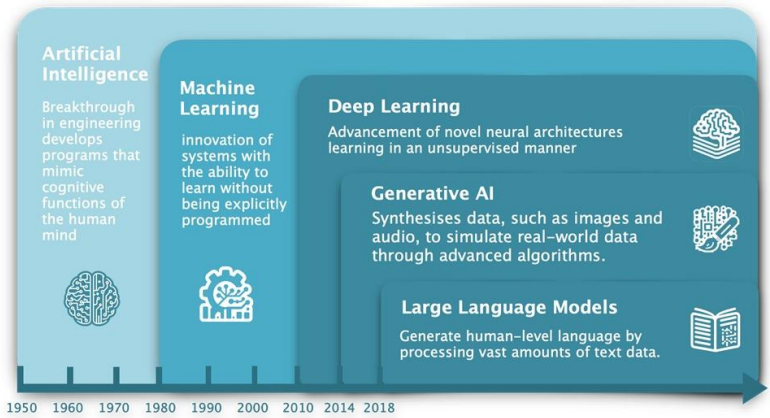

**Figure 2. Quantitative Metrics Evaluation of GPT-omni Responses**

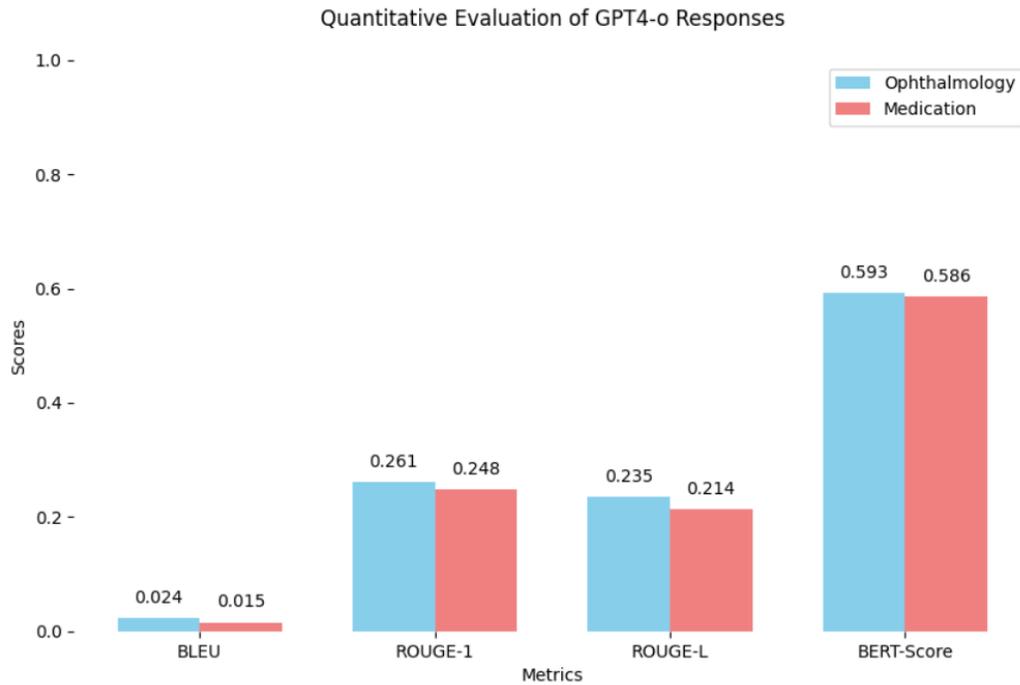

# Figure 3: SCORE Evaluation on Ophthalmology and Medication Queries

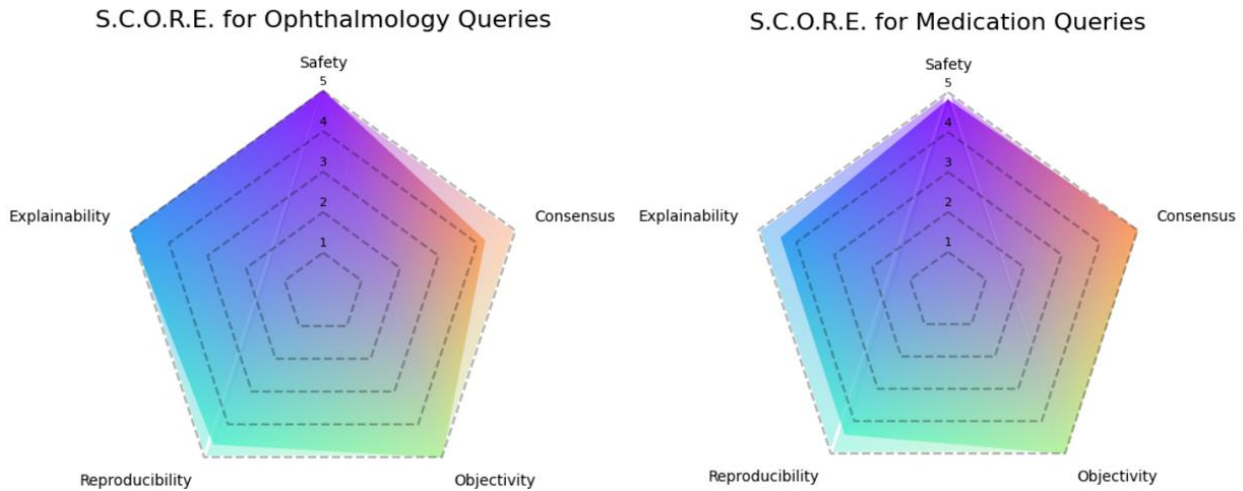

**Figure 4. Example Q&A pair for one of the medication-related questions.** The GPT4-omni generated response (GPT4-omni (1)) to the question is shown in the green table. Additional GPT4-omni (2) and (3) responses were generated to the same question for the assessment of 'Reproducibility' in S.C.O.R.E. The quantitative scores are shown in the yellow table, while the qualitative evaluation using the S.C.O.R.E. framework is shown in the blue table.

| | Question | Ground-truth |
|---|---|---|
| **Q&A pair** | Are there any genetic factors that influence azathioprine use? | Yes, some individuals may have an enzyme deficiency that can lead to severe toxicity when taking azathioprine. Genetic testing can help identify these individuals, and alternative therapies or dosage adjustments might be needed. |

| | GPT4-omni (1) | GPT4-omni (2) | GPT4-omni (3) |
|---|---|---|---|
| **GPT4-omni generated responses** | Yes, genetic factors can influence azathioprine use. Specifically, variations in the TPMT (thiopurine S-methyltransferase) and NUDT15 (nudix hydrolase 15) genes can affect how a person metabolizes azathioprine. Individuals with certain variants in these genes may be at higher risk for severe myelosuppression (bone marrow suppression) and may require dose adjustments or alternative therapies. Genetic testing can help guide safe and effective use of azathioprine. | Yes, genetic factors can influence the use of azathioprine. One key factor is the activity of the enzyme thiopurine S-methyltransferase (TPMT). Variations in the TPMT gene can affect how well your body metabolizes azathioprine. Individuals with low or absent TPMT activity are at higher risk for severe myelosuppression (bone marrow suppression) when taking azathioprine. Genetic testing for TPMT activity is often recommended before starting azathioprine therapy to guide dosing and reduce the risk of adverse effects. | Yes, genetic factors can significantly influence the use of azathioprine. The most notable factor is the activity of the enzyme thiopurine S-methyltransferase (TPMT). Variations in the TPMT gene can lead to different levels of enzyme activity:<br><br>1. **Low or Absent TPMT Activity**: Individuals with low or absent TPMT activity are at a higher risk for severe myelosuppression (bone marrow suppression) when taking azathioprine. This can lead to serious side effects such as infections and anemia.<br>2. **Intermediate TPMT Activity**: Individuals with intermediate TPMT activity may also be at risk for myelosuppression but to a lesser extent. They may require dose adjustments.<br>3. **Normal or High TPMT Activity**: Individuals with normal or high TPMT activity typically metabolize azathioprine more effectively and are less likely to experience severe myelosuppression.<br><br>Genetic testing for TPMT activity is often recommended before starting azathioprine therapy to guide dosing and minimize the risk of adverse effects. Another enzyme, nudix hydrolase 15 (NUDT15), can also influence azathioprine metabolism, particularly in certain populations, and may be considered in genetic testing. |

| | BLEU | ROUGE-1 | ROUGE-L | BERTScore |
|---|---|---|---|---|
| **Quantitative evaluation** | 0.049289012 | 0.348837205 | 0.279069763 | 0.6461799 |

| | S | C | O | R | E |
|---|---|---|---|---|---|
| **Qualitative evaluation** | 5 | 5 | 5 | 4 | 3 |

*Additional evaluation:*
*"Aligned with the ground truth in identifying the intended question (adverse drug reaction related to genetic polymorphism), aligning with evidence-based knowledge (genetic polymorphism increases the risk for severe toxicity, guideline-concordant actions) and reinforcing the need for genetic tests prior to initiation."*